\documentclass[conference]{IEEEtran}
\usepackage[utf8]{inputenc}
\usepackage[T1]{fontenc}
\IEEEoverridecommandlockouts

\usepackage{cite}
\usepackage{amsmath,amssymb,amsfonts}
\usepackage{algorithmic}
\usepackage{graphicx}
\usepackage{textcomp}
\usepackage{xcolor}
\usepackage{placeins}
\def\BibTeX{{\rm B\kern-.05em{\sc i\kern-.025em b}\kern-.08em
    T\kern-.1667em\lower.7ex\hbox{E}\kern-.125emX}}

\begin{document}

\title{Uydu Görüntülerinden Bina Bölütleme ve Son-İşleme Yöntemlerinin Performansı \\
 \huge Building Segmentation on Satellite Images and Performance of Post-Processing Methods
 {\footnotesize \textsuperscript{}} 
}

\author{\IEEEauthorblockN{Metehan Yalçın,Ahmet Alp Kındıroğlu,  Furkan Burak Bağcı, Ufuk Uyan, Mahiye Uluyağmur Öztürk}

\IEEEauthorblockA{\textit\textit{Huawei Turkey R\&D Center} \\
Istanbul, Turkey \\
\{metehan.yalcin,ahmet.alp.kindiroglu,  furkan.burak.bagci,ufuk.uyan1, mahiye.uluyagmur.ozturk\}@huawei.com
}}
\maketitle

\begin{abstract}

Researchers are doing intensive work on satellite images due to the information it contains with the development of computer vision algorithms and the ease of accessibility to satellite images. Building segmentation of satellite images can be used for many potential applications such as city, agricultural, and communication network planning. However, since no dataset exists for every region, the model trained in a region must gain generality. In this study, we trained several models in China and post-processing work was done on the best model selected among them. These models are evaluated in the Chicago region of the INRIA dataset. As can be seen from the results, although state-of-art results in this area have not been achieved, the results are promising. We aim to present our initial experimental results of a building segmentation from satellite images in this study.

\textit{Öz}--- Uydu görüntülerine erişilebilirliğin artması ve bilgisayarlı görü algoritmaların performanslarındaki artış ile birlikte araştırmacılar uydu görüntüleri üzerine de yoğun bir şekilde çalışmalar yapmaktadır. Uydu görüntülerinden bina bölütlemesi, şehir, tarım ve haberleşme ağlarının planlaması gibi birçok potansiyel uygulama için kullanılma potansiyaline sahiptir. Fakat, her bölge için edinilmiş uydu görüntülerinin olmayışı, herhangi bir bölgede eğitilmiş modelin genel özellikler kazanmasını önemli kılmaktadır. Bu çalışmada, Çin'de alınmış veri kümeleriyle modeller eğitilmiş, bu modeller arasından seçilen en iyi model üzerinde son-işlem yöntemleri uygulanmıştır. Bu modeller INRIA veri kümesinin Chicago bölgesinde test edilmiştir. Sonuçlardan da görüleceği üzere bu alandaki en iyi sonuçlar elde edilmemiş olsa bile sonuçlar ümit vericidir. Bu çalışmada, uydu görüntülerinden bina bölütleme deneylerinin ilk sonuçlarını paylaşılmak amaçlanmıştır. 
\textit{Anahtar Sözcükler}---bina bölütleme, uydu görüntüleri, denetimli semantik bölütleme, son-işlem, INRIA veri kümesi

\end{abstract}
\section{GİRİŞ}

Uydu görüntüleri üzerinden yeryüzü haritaları çıkartma uygulamaları hayatı kolaylaştırıcı işler yapmamızı sağlamaktadır. Şehir ve kırsal üzerinden alınmış görüntüler üzerinde çalışan bölütleme algoritmaları sayesinde yeryüzünde bulunan su havzaları, ormanlar, binalar ve daha birçok nesne tespit edilebilmektedir. Bu tespit yöntemleri sayesinde dağlık bölgelerdeki yollar ve yaşam alanlarını planlamak, kasaba ve köy gibi küçük yerleşim birimleri için yol, su, elektrik ve haberleşme ağlarının uygun noktalara yerleştirme işlemleri verimli bir şekilde gerçekleştirilebilmektedir. Bu sebeble uydu görüntüleri üzerinden yeryüzü bölütleme problemi büyük önem arzetmektedir. Çalışmamızda farklı bölgelerden alınan uydu görüntülerinden bina tespiti üzerine yoğunlaşılmıştır.  \\

Son yıllarda yapilan çalışmalarda anlamsal (semantik) bölütleme problemi üzerinde çokça mesafe katedilmiştir. Semantik bölütleme işlemi, görüntü üzerinde bulunan her piksel için bir sınıf seçiminin yapıldığı bir bilgisayarlı görü algoritmasıdır. Yarı-denetimli ve denetimsiz öğrenme yöntemlerinde etiketsiz görüntülerden faydalanılabilirken, denetimli semantik bölütleme eğitiminde uygun bir sekilde etiketlenmiş görüntülere ihtiyaç duyulmaktadır. Uydu görüntüsü üzerinde bulunan her piksel için o pikselin hangi sınıfa ait olduğunu belirten etiket dosyaları bulunmaktadır. Bu çalışmada da bina bölütleme problemi üzerine çalışılmakta, veri kümesindeki bina sınıfları kullanılmaktadır. Her görüntünün içerisindeki binalara ait pikselleri ifade eden piksellere etiket dosyasında 1, binayı ifade etmeyen piksellere ise 0 değeri verilmiştir. \\ 

Burada karşılaşılan zorluklardan biri modelleri eğittiğimiz görüntüler ile test bölgesi uydu görüntülerinin yakınlaştırma seviyelerinin farklı olabilmesidir. Bu çalışmada farklı yakınlaştırma düzeylerinden alınmış görüntülerle eğitilmiş modellerin, INRIA\cite{maggiori2017dataset} veri kümesi üzerindeki performansı incelenmiştir. \\

Semantik bölütleme modeli eğitimi için gerekli verilerin çokluğu, modelin genelleştirilebilmesi adına önem arz etmektedir. Eğitim verisi sayısının artması modelin daha farklı görüntüler görmesine ve öğrenmesine yol açmaktadır. Bu çalışmada Çin'in bina oranı yüksek bölgelerinden alınmış görüntülerle denetimli semantik bölütleme modelleri eğitilmiş ve aynı model INRIA veri kümesi icerisinde bulunan "Chicago" şehrinde test edilmiştir. Bahsettiğimiz gibi denetimli öğrenme yönteminde veri sayısının çokluğu işe yarıyorken, veri kümesinin içerisinde bulunan veri dağılımlarının benzerliği de önemli olduğu görülmüştür. Binaların yoğun olarak bulunduğu bir şehir bölgesi ile tarımsal arazi bölgesi bir arada eğitildiğinde modele bağlı olarak öğrenememe sorunları yaşanmıştır. Alan farklılığından yaşanan bu sorunlar öğrenme transferi \cite{druck-etal-2009-active} yöntemleriyle çözülebilmektedir. Bu çalışmada sadece bina sıklığının yoğun olduğu bölgelerle eğitim yapılıp test edilmiştir. \\

\section{Literatür Özeti}
Görüntüler üzerinde sınıflandırma, obje bulma, anlamlandırma (image-captioning), görüntüler arası transfer ve yeni görüntüler üretme gibi işlemlerde başarılı pek çok bilgisayarlı görü yöntemi vardır. Semantik bölütleme de bilgisayarlı görünün önemli kullanım alanlarından birisidir. Semantik bölütleme işleminde piksel seviyesinde tahminler yapılmaktadır. Zaman içerisinde semantik bölütleme için birçok yöntem önerilmiştir. FCN\cite{long2015fully} semantik bölütleme problemine uygulanan ilk çözümlerden biridir. FCN mimarisinde birçok evrişimsel ve havuzlama (pooling) katmanla görüntünün boyutu 1/32 katına küçültülmüş ve ters-evrişim katmanları ile görüntü gerçek boyutuna getirilerek piksel bazında tahminler yapılmıştır. \\

Bahsetmemiz gereken bir diğer önemli mimari ise U-Net\cite{ronneberger2015u} modelidir. Bu calışmada U-Net'in geliştirilmiş versiyonu deneylerde kullanılmaktadır. U-Net mimarisinde, FCN e gore iki önemli değişiklik göze çarpmaktadır. FCN mimarisi sırasıyla aşağı ve yukarı örnekleme yaparak gerçek görüntü çözünürlüğünü elde etmekte olup, U-Net yapısı bunu kodlayıcı ve kod çözücü yapılarını simetrik olarak kullanarak yapmakta ve kod çözücü yapısında çoklu kanal yapısı kullanarak kodlayıcı kanalının katmanlarındaki semantik bilgiyi kod çözücü kanalın ilgili kanallarına aktarabilmeyi başarmıştır. Bu şekilde son katmandaki semantik bilgi ilk katmandakilerle birlikte güçlendirilmiştir. U-Net++ \cite{zhou2018unet++} çalışması ile bu yapı iyileştirilmiş, kodlayıcı ve kod çözücü kısımları arasında atlama-bağlantısı (skip-connection) yapısı kullanılmıştır. Atlama-bağlantısı yapısı ile ilk katmanlardaki semantik bilginin ileriki yukarı-örnekleme katmanlarına aktarımı sağlanarak güçlendirilmesi amaçlanmaktadır. Günümüzde görüldüğü üzere medikal görüntülerde kullanılmak icin tasarlanan bu mimari, uzaktan algılama dahil birçok alanda önemli başarımlar elde etmiştir. Bu calışmada U-Net++ modelinden faydalanılarak eğitilen modeller, INRIA veri kümesinde test edilmiştir. \\

DeepLab \cite{chen2014semantic} mimarisi de yukarıda bahsettiklerimiz gibi evrişimsel ağları kullanan bir başka yöntemdir. Bu yöntemde birkaç yeni özellik eklenmiştir. Delikli (atrous) evrişim diger ismiyle genişletilmiş (dilated) evrişim denilen yeni bir yöntemle daha büyük bir görüntü alanı, daha az parametre kullanarak evrişim işlemi yapılmış ve işlem yükü azaltılmıştır. DeepLabV1 geliştirmesinde ise aşağı-örnekleme ile kaybedilen uzamsal çözünürlük daha az aşağı-örnekleme yapılarak bunun yerine delikli evrişimsel filtrelerinin büyütülmesi ile çözülmüş, aynı zamanda tamamen bağlı koşullu rasgele alan(fully connected conditional random field ) ile benzer pikseller kullanılarak sınıf tahminine katkıda bulunmuşlardır. DeepLabV2 \cite{chen2017deeplab}, DeepLabV3 \cite{chen2017rethinking} ve son olarak bu çalışmada da kullanılan DeepLabV3+\cite{chen2018encoder} yöntemlerinde de Delikli uzamsal piramit (Atrous Spatial Pyramid) havuzlama ile farklı örnekleme sıklığı ile elde edilmiş çıktıların birleştirilmesiyle sonuçlar iyileştirilmiş, daha keskin nesnelerin kestirimi icin delikli evrişim işlemini U-Net'de bulunan kodlayıcı ve kod çözücü mimarisine adapte etmişlerdir. Bu sayede son modelde daha keskin sınırların bulunulduğu ve semantik bilgiye daha iyi bir performansla erişilebildiği gözlenmiştir. Li ve arkadaşlarının \cite{DeeplabALIL} calışmasında DeeplabV3+ \cite{chen2018encoder} modelini bina bölütleme icin kullanmış, ayrıca bu model sonuçlarını iyileştirmek için aktif öğrenme, artımlı öğrenme ve  transfer öğrenimi yöntemlerinin etkilerini gözlemlemiştir. \\

Uydu görüntülerinden bina bölütleme problemi yakın süre içerisinde birçok araştırmacının ilgisini çekmektedir. Bu çalışmalarda derin öğrenme metodlarını kullananlar bulunduğu gibi, destek vektör makinesi (SVM) metodunu kullanarak bina bölütleme yapıldığı çalışmalar da bulunmaktadır\cite{vakalopoulou}. Vakalopoulou ve arkadaşlarının çalışmasında INRIA veri kümesinden alınmış yüksek cözünürlüklü uydu görüntüsünü önce AlexNet \cite{ALEXNET}'e beslemiş, FC (Fully Connected) yapısının 7. katman çıktısını SVM sınıflandırıcısını eğitmek icin özellik vektörü olarak kullanmışlardır. Eğitilen SVM\cite{cortes1995support} modeli çıktısı MRF (Markov Random Field) tabanlı modelle rötuş (post processing) işlemi yapılarak sonuçlar iyileştirimiştir. Fakat bina bölütleme problemi görüntüler arasında ölçek faktörü farklılığı, gölge oranı, binaya benzeyip bina olmayan komplex objelerin varlığı sebebiyle klasik görüntü işleme yöntemleriyle çözülemeyecek kadar zor bir problemdir. Bu sebeble derin öğrenme algoritmaları kullanılarak yakın zamanda pek çok calışma gerçekleştirilmiştir. \\

Yang ve arkadaşlarının \cite{Yangetal} yaptığı calışmada farklı seviyedeki özellik çıktılarını kullanan dikkat (attention) mekanizmasını kullanan yeni bir mimari DenseNets ile bina bölütleme calışmaları yapılmıştır. Bittner ve arkadaşları \cite{8447548} ise yukarıda bahsettiğimiz FCN \cite{long2015fully} yapısını kullanarak yüksek çözünürlüklü uydu görüntüleri ile aynı görüntünün yükseklik bilgilerini taşıyan sayısal yeryüzü modeli (DSM) verilerini kullanarak bina bölütlemesi yaptı. Marcu ve arkadaşları \cite{multistage} çok-aşamalı çoklu-görev mimarisi ile bir dalında bölütleme yaparken diğer regresyon dalında lokalizasyon yaparak çoklu dal kullanımının bölütleme sonuçlarını iyileştirdiğini gösterdi. \\

Uzaktan algılama görüntülerinde bina bölütleme ile ilgili birçok karşılastırma makalesi de bulunmaktadır \cite{Hu2019},\cite{Yietal},\cite{Erdem},\cite{comparisonChina}. 
Yi ve arkadaşlarının \cite{Yietal} yaptığı çalışmada U-Net modeline göre daha az sonuç çıkarma süresine sahip, az parametreyle daha iyi sonuçlar veren DeepResUnet modelini geliştirdi ve bu modeli DeepConvNet\cite{DeepConvNet}, FCN, UNET , Segnet\cite{SegNet} modellerini kullanarak bina bölütleme performansları karşılaştırması yaptı. Erdem ve diğerleri \cite{Erdem} bina bölütlemesine uygun model bulabilmek icin U-Net modelini birkaç farklı öznitelik çıkarıcı mimarileriyle eğiterek performanslarını karşılaştırdı, aynı zamanda da çoğunluk oylama son işlem yaklaşımı uygulayarak sonuçların iyileşip iyileşmediğini test etti. \\

Ayrıca bina bölütleme modellerinin performansını geliştirmek icin kullanılabilecek veri arttırım yöntemlerinin karşılaştırıldığı calışmalar da bulunmaktadir \cite{rs14092012}, \cite{9607556}. Illarionova ve arkadaşları \cite{9607556} eğitim veri kümesini arttırabilmek için yeni bir obje tabanlı veri arttırım yöntemi geliştirdi. Bu yöntemde Hedef piksellerini ile temiz arka plan görüntülerine sahip görüntülere aktararak elde bulunan hedef pikselli görüntü sayısını arttırdı. Bu metodu birçok bölgede test ederek tutarlılığını kontrol ettiler. Li ve arkadaşlarının çalışmasında ise \cite{jasonpostprocessing}2000 li yıllardan bu yana yapılmış bütün bina bölütleme ve tespit çalışmalarını araştırmı ve makalelerinde veri arttırım yöntemleriyle birlkte sunmuşlardır. Bu çalışmaya göre bina tespitleri, önceden edinilmiş geometrik, spektral ve içerik bilgileri ile yapılan son-işlem yöntemleri ile iyileştirilebilmiştir \cite{Huang2017}. Spektral bilgi varsa NFBI  (normalized difference vegatiation index - NDVI) veya HSV renk uzayındaki renk tonu bilgisi kullanarak toprak bölgelerdeki yanlış alarmlar temizlenmiştir. Geometrik analiz kısmında ise tahminler içerisinde bağlı bileşen analizi yapmışlardır. Son olarak da içerik bilgisi ile son-işlem yaparken sahnedeki gölgeler tepit edimiş, gölge bulunmayan yerdeki bina tespitleri elimine edilmiştir.\\

INRIA veri kümesi ABD ve Avusturya ülkelerinin birçok şehrinden alınmış yüksek cözünürlüklü uydu görüntülerinden oluşmuş, bina sınıflarının etiketlerinin bulunduğu bir veri kümesidir. Çalışmamızda bu veri kümesinin Chicago bölgesinde test yapılmıştır. Eğitim veri kümesi ise Çin üzerinden alınmış yüksek çözünürlüklü, bina yoğunluğunun sık olduğu alanlardan oluşmaktadır. HWLC-18 ve HWLC-16 veri kümesi Huawei tarafından oluşturulmuştur \cite{ubmk}. Bu veri setleri sırasıyla 0.6 ve 2.4 metre çözünürlüğe sahip olup openstreetmap'e göre seviyeleri Seviye-18 ve Seviye-16 olarak kabül edilmektedir. \\

Bu makalede U-Net++ ve DeepLabv3+ mimarilerini, farklı kodlayıcı modelleri kullanarak eğitimler düzenlenmiş ve modeller oluşturulmuştur. VGG-16 \cite{7486599}, EfficientNet\cite{pmlr-v97-tan19a}, Resnet \cite{7780459} and MobileNet \cite{mobilenet} yapılarından öznitelik çıkarıcı olarak faydalanılmıştır. Test adımı sırasında uygulanan son-işlem yöntemlerinin model performasına etkileri incelenmiştir.  \\

\section{Yöntem}

Uydu görüntüleri üzerinden bina tespiti yapmak amacıyla bilgisayarlı görü uygulamaları sıkça kullanılmaktadır. Bu çalışmada temel olarak U-Net++ ve DeepLabV3+ modeli temel alınmıştır. Bu modeller farklı öznitelik çıkarıcı ağlar kullanılarak eğitilmiş ve sonuçları karşılaştırılmıştır.  \\

\begin{figure}[htbp]
\centerline{\includegraphics[width=1.0\columnwidth]{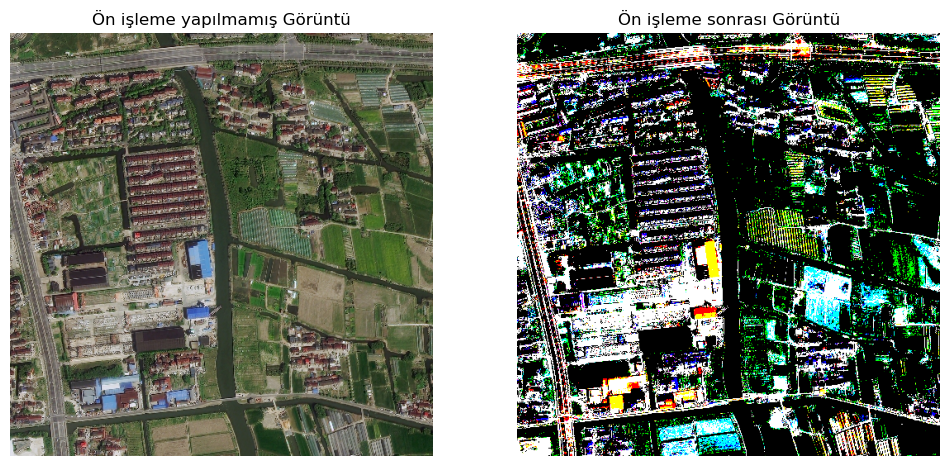}}
\caption{Görüntü Ön İşleme Adımı}
\label{fig:onislemee}
\end{figure}

Kentsel olarak gelişmis bir bölgeden alınan uydu görüntüleri geniş bir alan kaplamaktadır. Uzamsal olarak yüksek çözünürlüklü bu resimler model eğitiminde kullanabilmek icin kırpılmıştır. 512x512 boyutunda kırpılmış kırmızı-yeşil-mavi (RGB) kanallı resimler ile yine aynı boyuttaki etiket dosyaları oluşturulmuştur. Bu etiket dosyaları bina ve bina olmayan sınıfları olarak iki farklı sınıf etiketinden oluşmaktadır. Eğitim sırasında veri çeşitliliğinin arttırılması için veri arttırımı görüntü işleme metodları kullanılmaktadır. Kullandığımız öznitelik çıkarıcı mimarileri en iyi ağırlıklarını Imagenet \cite{ImageNet} veri kümesi üzerinde eğitilmiş ve bu veri kümesinin ortalama ve standart sapma değerlerini kullanmışlardır. Bu sebeble eğitimin ilk basamağında görüntüler ön-işleme yapılarak uydu görüntüleri ortalama ve standart sapma değerleri Imagenet ile aynı seviyeye getirilmiştir. Şekil \ref{fig:onislemee} de Resnet mimarisi ön-işleme adımı öncesi ve sonrası görüntü gösterilmiştir.\\

Denetimli semantik bölütleme yöntemlerinde veri çesitliliği ve sayıca fazlalığı modelin genel öznitelikler öğreniminde önemli yer kapsamaktadır. Yüksek çözünürlüklü görüntülerden kırpılarak elde ettiğimiz eğitim veri kümesi her eğitim basamağında grup boyutunca (batch size) çeşitli veri arttırım yöntemleri uygulanmaktadır. Bu veri arttırım yöntemi seçimi rastgele yapılmaktadır. Çubuk ve arkadaşlarının \cite{cubuk2020randaugment} çalışmasında veri arttırım yönteminin semantik bölütleme üzerindeki etkisini ortaya koymuşlardır. Çalışmamızda da benzer bir yöntemden faydalanılmış ve veri arttırımı sağlanmıştır. Eğitim sırasında grup boyutu 8 olarak belirlenmiştir. Grup içerisindeki her görüntü için liste halinde bulunan veri arttırım yöntemlerinden rastgele biri uygulanmış, uygulanıp uygulanmayacağı da bir olasılık yüzdesine bağlanmıştır. Döndürme, Afin dönüşüm, Transformasyon, Görüntü Renk Değerleri Tersine Çevirme, Rastgele Kontrast, Rastgele Parlaklık ekleme yöntemleri  \textit{albumentations} \cite{albumentations} kütüphanesi kulanılılarak uygulanmış, eğitim basamak sayısı arttıkça döndürme ve diğer transformasyonların derecesi yükseltilmiş ve daha etkili uygulanması sağlanmıştır. \\

Bu çalışmada temel olarak U-Net++ ve DeepLabV3 modelleri kullanılmaktadır. U-Net++ mimarisi isminden de anlaşıldığı üzere U harfi şeklinde evrişimsel sinir ağlarından oluşmaktadır, bu yapı otomatik-kodlayıcı (auto-encoder) yapılarından esinlenilerek tasarlanmıştır. U-Net++ yapısının ilk versiyonlarına teme farkı atlama-bağlantısı (skip-connection) yapısı kullanarak semantik bilginin bütün katmanlar arasında yayılımını sağlamaktadır. Farklı öznitelik çıkarıcılar kullanılarak U-Net++ modeli performans değerlendirmesi yapılmıştır. Bunlardan ilki VGG-16 mimarisidir. VGG-16 bir çeşit evrişimsel sinir ağı (CNN) dır ve ağırlık içeren 13 adet evrişim katmanı ve 3 adet yoğun sinir ağından oluşmasından dolayı VGG-16 ismini almıştır. VGG-16 yapısını diğer öznitelik çıkarıcılardan ayıran özelliği, bütün katmanlarında 3x3 evrişim ağı bulundurmasıdır. Daha küçük  ve sabit 3x3 boyutu evrişim filtreleri kullanmaları AlexNet'e göre daha az sayıda parametre kullanmalarına aynı zamanda daha iyi performans kazanmaları ile sonuçlanmıştır. Yapısında 5 evrişim katmanı bulunmakta bu katmanlar arasında da havuzlama yapılmaktadır. Son katmanda ise 3 adet yoğun sinir katmanı (Dense layer) kullanarak sınıflandırma problemlerinde ullanılmak üzere üretilmiştir. Çalışmada öznitelik çıkarıcı olarak kullanıldığı için son katman kullanılmamakta, ara katmanların ürettiği aktivasyonlar öznitelik olarak kullanılmaktadır. \\

Modellerdeki parametre çokluğu işlemsel yük getirmektedir. Daha düşük parametreliş modeller elde etmek için boyutlar azaltılıyor, bu halde öznitelikleri kaybetmemek için çözünürlük ve genişlik büyütülmeye çalışılıyor fakat her mimaride bu ayrı ayrı yapılabilmektedir. EfficientNet yapısında ise daha az parametre ile hem derinlik hem genişlik hem de çözünürlükte artış sağlayıp verimli bri şekilde tespit yapılabilmesini sağlayan bir yapı geliştirilmiştir. Bunu gerçekleştirebilmek için hem derinlik hemde nokta tabanlı evrişim yapılmaktadır. Bu yönteme birleşik katsayı (compound coefficient) yöntemi denilmektedir. \\

ResNet mimarisi de çalışmamızda kullanılan bir diğer önemli öznitelik çıkarıcı yapıdır. Bu yapının temel çıkış noktası VGG-16 gibi evrişimli nöral ağ yapılarındaki gradyan tükenmesi (vanishing gradient) probleminin önüne geçmektir. Bu sebeble atlama-bağlantısı (skip-connection) yapılarını kullanarak katmanlar arası bağlatı sağlayarak gradyanların sıfırlanma problemine çözüm getirmişlerdir.\\

\begin{figure}[htbp]
\centerline{\includegraphics[width=1.0\columnwidth]{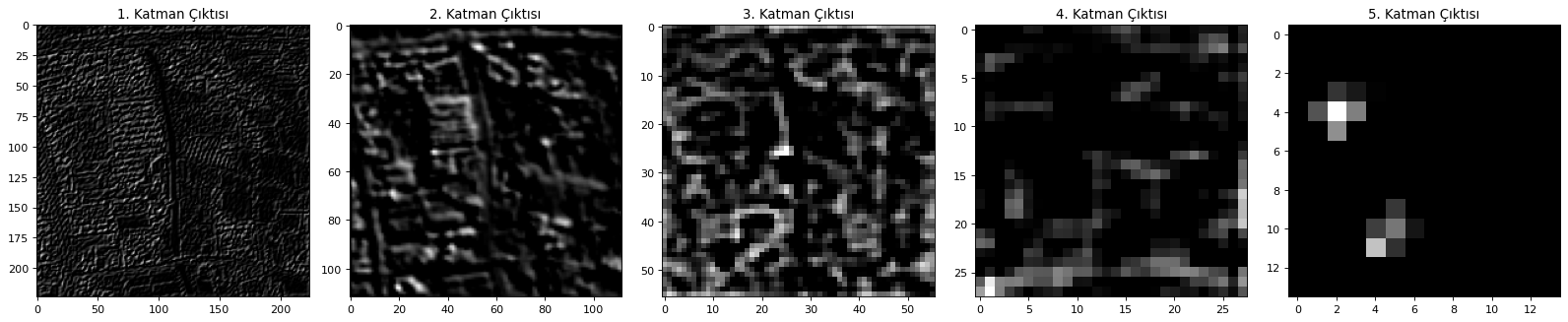}}
\caption{VGG-16 öznitelik çıkarıcısı katman çıktıları}
\label{fig:vggkatmanlar}
\end{figure}

\begin{figure}[htbp]
\centerline{\includegraphics[width=1.0\columnwidth]{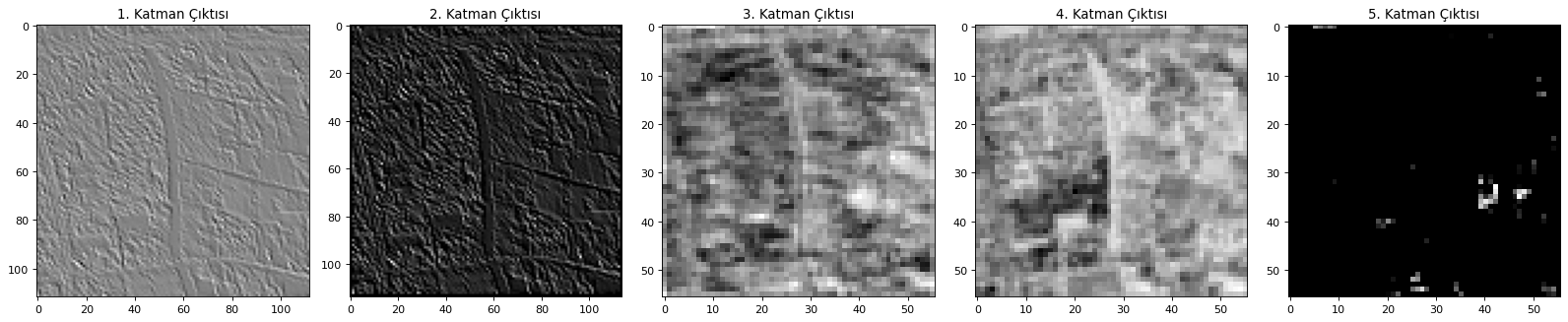}}
\caption{Resnet-50 öznitelik çıkarıcısı katman çıktıları}
\label{fig:resnetkatmanlar}
\end{figure}

\begin{figure}[htbp]
\centerline{\includegraphics[width=1.0\columnwidth]{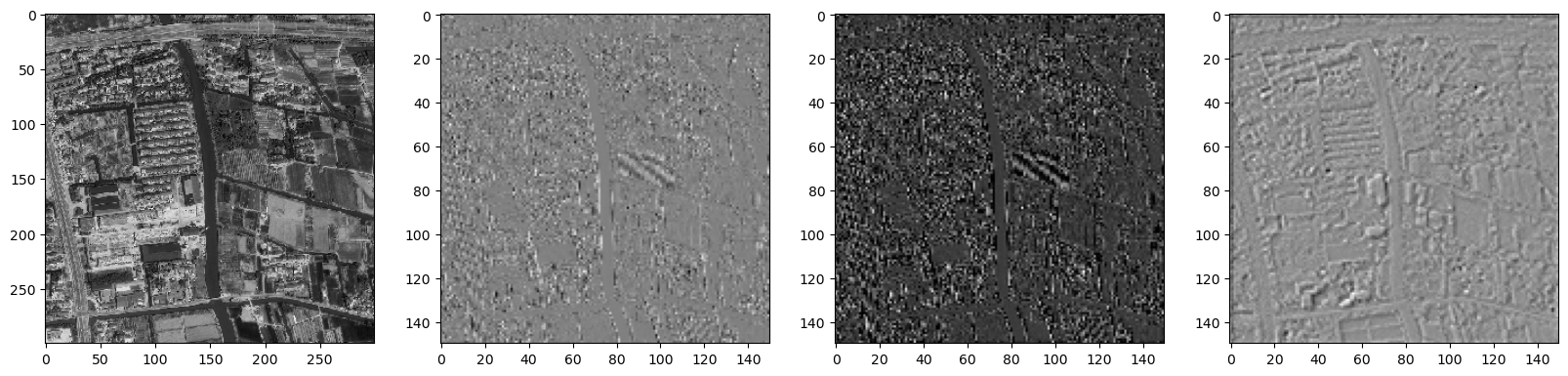}}
\caption{EfficientNet-B0 öznitelik çıkarıcısı katman çıktıları}
\label{fig:efficientnetkatmanlar}
\end{figure}

Öznitelik çıkarıcı mimarilerin performansını değerlendirebilmek için uydu görüntüleri bu yapıların girişine vererek, içerdikleri katmanların çıktıları görselleştirilmiştir. Şekil \ref{fig:vggkatmanlar} de VGG-16 öznitelik çıkarıcısının sırasıyla ilk 5 katman çıktısı görselleştirilmiş, Şekil \ref{fig:resnetkatmanlar} de Resnet-18 ve Şekil \ref{fig:efficientnetkatmanlar} aktivasyonları görselleştirilmiştir. Buradan da görüldüğü üzere VGG-16 modeli bina yüzeylerini daha belirgin şekilde ilk katmanlarında bulabilmekte ve bunu 
son katmanlara aktarabilmektedir. EfficientNet modeli keskin hat bulma konusunda VGG-18 den geri kalmasına krşın ResNet-50 ye göre daha küçük detaylı ortaya çıkarabilmiştir. \\

ResNet-50 modelinin alıcı alanı (receptive field) 483, VGG-16 ağının da 212 dir. Alıcı Ağının gördüğü alanın darlığı bu problemde daha iyi performans sağladığı düşünülebilir. 





Bu calışmada 3 farklı kayıp fonksiyonu denemesi yapılmıştır. Bunlardan ilki, 
\subsection{Dice Kayıp}
\textit{Dice} kayıp fonksiyonu medikal görüntü bölütleme problemlerinde yaygın olarak, veri dengesizliği bulunan eğitim zorluklarında sıkça kullanılan, \textit{Dice} katsayısından esinlenilerek oluşturulmuş bir kayıp fonksiyonudur. Çapraz Düzensizlik (Cross Entrophy) fonksiyonuna sıkça tercih edilme sebebi veri içerisindeki sınıf dağılımının dengesiz olmasından dolayıdır. 

\subsection{Focal Tversky Kaybı}
Tversky Kaybı \cite{TL} dice kaybında da bulunan DP (doğru pozitif) ve YN (yanlış negatif) terimlerine ağırlıklandırma verilmesi ile elde edilmiştir. Beta terimi 0,5 olarak seçildiğinde Dice kaybı ile aynı duruma gelmektedir. Focal Tversky kaybı'nda ise, zor örnekleri çözümleyebilmek icin bazı katsayılar eklemiştir \cite{FTL}. 

\subsection{Ağırlıklandırılmış Dice Kaybı}*
Ağırlıklandırılmış Dice kaybı ise Dice kaybını çoklu etiket şeklinde kullanırken her sınıfın kayba etkisini değiştirir. Bu şekilde modelin belirli bir sınıfı daha iyi öğrenmesini sağlayarak kaybın azaltmasına yol açar.

Bu çalışmada test sırası görüntü arttırımı yöntemleri kullanılarak test zamanı bölütleme performansı testleri yapılmıştır. Bu kütüphane içerisinde \textit{yatay döndürme}, \textit{dikey döndürme}, \textit{istenilen açıda döndürme}, \textit{çözünürlük değiştirme} gibi birçok görüntü manipülasyon teknikleri bulunmaktadır. Son-İşleme adımındaki temel mantık görüntüyü farklı açılar ve durumlarda test ederek bu test sonuçlarının birleştirilmesi ve sonuca etkisinin gözlenmesidir. Bu calışmada farklı test zamanı veri arttırımı yöntemleri denenerek sonuçları Tablo\ref{tab:tab6} de görüldüğü gibi eklenmiştir. Yöntem-1 de yapılan test zamanı veri arttırımında \textit{Çarpma}, \textit{yatay döndürme}, \textit{çözünürlük değiştirme}, ve \textit{istenilen açıda döndürme} işlemleri uygulanmıştır. Bütün yapılan veri arttırımlarını \ref{tab:tab6} da görebilirsiniz. \textit{İstenilen açıda döndürme} işlemi görüntüyü belirlenen derecelerde saat yönünde döndürme işlemidir. \textit{Çözünürlük değiştirme} işleminde ise görüntü belirlenen değerlerde büyütülüp küçültülmektedir. \textit{Çarpma}da ise görüntü üzerindeki her piksel bu değerler ile çarpılarak yeni görüntüler elde edilmektedir. Son olarak \textit{yatay döndürme}de ise görüntü x ekseni doğrultusunda 90 derece döndürülmektedir.

\begin{table}[htbp]
\caption{Son-İşleme Yöntemleri}
\begin{center}
\resizebox{.99\hsize}{!}{$
\begin{tabular}{|c|c|c|c|c|}
\hline
\textbf{Yöntemin Adı} 	& \textbf{Yatay Döndürme} 	& \textbf{İstenilen Açıda Döndürme}	& \textbf{Çözünürlük Değiştirme} 		& \textbf{Çarpma} 	\\ \hline
Yöntem-1       				& +           					& 0, 180  			& [1]   					& [0.9,1,1.1]              \\ \hline
Yöntem-2       				& +           					& 0, 180         		& [0.25,0.5,0.75,1]      & [0.9,1,1.1]       		\\ \hline
Yöntem-3        				& +        					& 90         			& [0.5,0.75,1] 			& -                     \\ \hline
\end{tabular}\label{tab:tab6}$}
\end{center}
\end{table}

\begin{figure}[htbp]
\centerline{\includegraphics[width=1.0\columnwidth]{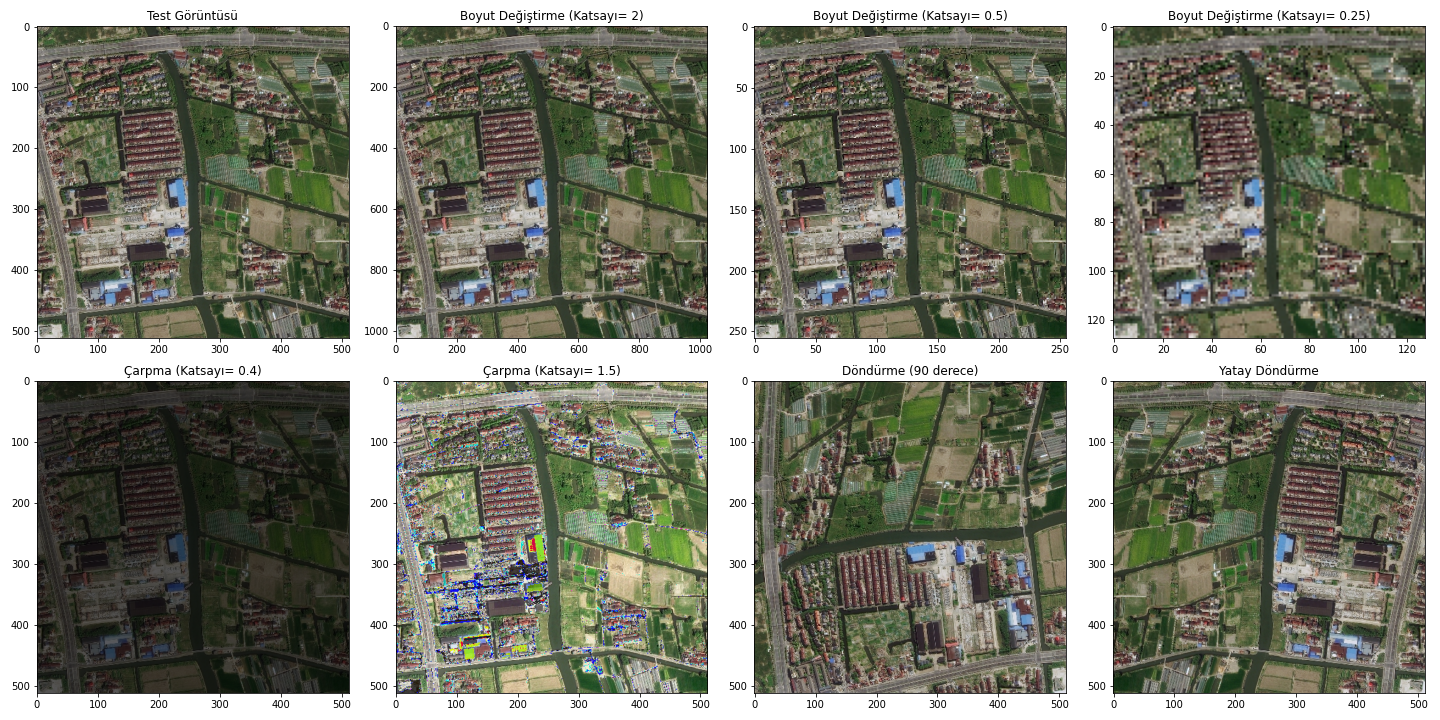}}

\caption{Test Zamanı Veri Arttırım İşlemleri Örnekleri}
\label{fig:tta}
\end{figure}

 

\section{Uygulama Detayları, Değerlendirme Kriterleri ve Deneyler}\label{Aa}
\subsection{VERİ KÜMESİ}
Eğitim veri kümesi olarak Huawei tarafından Çin'in şehirleşmenin yoğun olduğu bölgelerinden alınan uydu görüntüleri eğitim setimizi oluşturmaktadır. HWLC-16 ve HWLC-18 veri kümeleri sırasıyla 2.4 ve 0.6 metre çözünülüge sahip olmakta olup test setinde ullanılan INRIA veri kümesinin Chicago verilerinin çözünürlüğünden (0.3 metre) farklıdır. Uydu görüntüleri ile ilgili önemli kriterlerden biri yakınlaştırma seviyeleridir. Bu çalışmada yapılan eğitimlerde Seviye 18 ve Seviye 16 veri kümeleri kullanılmıştır. Bu veri kümeleri sırasıyla piksel başına 0.6-metre ve 2.4-metre çözünürlüğe sahiptir. Görüntüler içerisinde bulunan objeler 21 sınıf olacak şekilde etiklenmişlerdir. Çalışmamızda sadece bina etiketleri kullanılmış ve diğer cisimler bina-değil olarak kabul edilmiştir. Ayrıca modelin genel öznitelikler öğrenip öğrenmediğini test etmek için INRIA veri kümesinde bina bölütleme yapılmaya çalışılmıştır. Amerika'nın çeşitli şehirlerinden alınmış görüntüler 0.3-metre çözünürlüğe sahiptir. Bu çalışmada sadece INRIA veri kümesinin "Chicago" şehri baz alınmıştır. 

\subsection{Uygulama Detayları}
Yapılan deneylerde bazı parametreler farklılık göstermekle birlikte ortak tutulan değerler de bulunmaktadır. Deneylerde grup boyutu 8, öğrenme hızı 0.0001 olarak ayarlanmış, optimizasyon yöntemi olarak da ADAM\cite{kingma2014method} baz alınmıştır. Bunların haricinde kayıp fonksiyonu denemeleri neticesinde her deney için ağırlıklandırılmış dice kayıp fonksiyonu kullanılmıştır. Bölütleme proleminde sıkça kullanılan metrik olan kesiştirilmiş bölgeler ölçütleri (intersection over union, IoU) metriği ile değerlendirilme yapılmıştır. Jaccard Index yani  IOU uygulanarak bina sınıfına ait IoU ve ortalama IoU(mIoU) değeri hesaplanmıştır. Eğitimlerimiz  Geforce Nvidia 2080TI kartı kullanılarak yapılmış, her model en fazla 20 basamak olacak şekilde eğitilmiştir.

\subsection{Değerlendirme Kriteri}
Değerlendirme metriği olarak IoU (Intersection over Union) kullanılmıştır. Doğruluk hesabı yapılırken IoU olarak bilinen Jaccard Index'ten faydalanılmıştır. Bu index, Denklem \ref{eq:4}'de görüldüğü gibi hesaplanmaktadır. DP (Doğru Pozitif), YP (Yanlış Pozitif), DN (Doğru Negatif) değerleri sırasıyla hedefin doğru tahmin edildiği, hedef olduğu yönünde tahmin yapılıp yanlış olan, ve hedef olmadığı yönünde tahmin yapılıp aslında hedef olan piksel durumlarını ifade etmektedir. Bunların oranı bize hedef olarak tahmin ettiğimiz bölgenin aslında ne kadarının doğru tespit edildiğini belirtir.

\begin{equation}
\resizebox{.4\hsize}{!}{$IoU = \frac{DP}{DP+YP+DN}$} \label{eq:4}
\end{equation} 

\subsection{Deneyler}
Bu çalışmanin temel hedeflerinden biri bina tespiti yapmakken beraberinde modellerin başka ülke ve bölgelerdeki performansları da incelenmek istenmiştir. Eğitimler Çin de bulunan yoğun bina iceren şehir bölgelerinden alınmış, 0.5-metre ve 2-metre çözünürlükte görüntülerle yapılmış, bu modeller ise INRIA veri kümesinin Chicago'dan alınmış 0.3-metre çözünürlükle elde edilen görüntülerde test edilmiştir. 

Deneylerin ilk kısmında U-Net++ ve DeepLabV3+ modeli tablo \ref{tab1} de görüldüğü gibi 0.6 ve 2.4-metre çözünürlükteki verilerle eğitilmis modellerin hedef bolgedeki test sonuçlari verilmiştir. 

\begin{table}[htbp]
\caption{Farklı Derin Öğrenme mimarileri kullanılarak eğitilmiş modellerin INRIA veri kümesindeki performans tablosu. }
\begin{center}
\resizebox{.99\hsize}{!}{$
\begin{tabular}{|c|c|c|c|c|c|}
\hline
\textbf{Model} 	& \textbf{Öznitelik Çıkarıcı} 	& \textbf{mIoU}	& \textbf{Bina IoU} 		& \textbf{Trained Zoom Level} 	 \\ \hline 
Unet++       		& VGG-16           		& \textbf{0.701}  &\textbf{0.811}   		                      			&    HWLC-18 (0.6m)				 \\ \hline 
Unet++       		& VGG-16           		& 0.582         	&0.716          			                  			& 	HWLC-16 (2.4m)				 \\ \hline 
Unet++       		& EfficientNet-B3       & 0.642         	&0.765          		                     			& 	HWLC-18 (0.6m)				 \\ \hline 
DeepLabV3++	& EfficientNet-B5       & 0.569         	&0.709          			                     			& 	HWLC-18 (0.6m)			 	 \\ \hline
DeepLabV3++	& EfficientNet-B3       & 0.491         	&0.206          			              			&    HWLC-16 (2.4m)				 \\ \hline
DeepLabV3++    &  Resnet50           	& 0.621         	&0.443          			       			& 	HWLC-16 (2.4m)				 \\ \hline
DeepLabV3++    &  Resnet18           	& 0.644         	&0.489          			               			&    HWLC-16 (2.4m)				 \\ \hline
DeepLabV3++    &  Resnet34           	& 0.631         	&0.502          			             			& 	HWLC-16 (2.4m)				 \\ \hline
DeepLabV3++    &  Mobilenetv3           	& 0.576         	&0.452          		            			& 	HWLC-16 (2.4m)		 \\ \hline
\end{tabular}\label{tab:tab1}$}
\end{center}
\end{table}

Tablo \ref{tab:tab1} üzerinde yapılan deneyler görülmektedir. Deneylerde görüldüğü üzere U-Net++ mimarisi ile VGG-16 öznitelik çıkarıcı en iyi sonuçları vermektedir. U-Net++ ile yapilan farklı yakınlaştırma seviyeleri denemelerinden görülecegi üzere yüksek çözünürlüğe sahip 0.6-metre çözünürlüklü verilerle yapılan eğitimlerde 2.4 ile yapılan verilerle yapılana göre daha iyi sonuçlar elde edilmiştir. Benzer çözünürlüğe veya daha yüksek çözünürlüğe sahip bir veri kümesinin hiç görmediği benzer karakteristiğe sahip bölgelerde de başarılı olabileceğini görmüş bulunmaktayız. Buradaki deneylerden bir diğer çıkarımımız da U-Net++ modelinin DeepLabV3+ dan daha olumlu sonuçlar vermesidir. \\

DeepLabv3+ metodu Resnet-18 öznitelik çıkarıcı ile en iyi sonuçları verdiği görülmektedir. DeepLabV3+ modelini Resnet-18 öznitelik çıkarıcı ile kullanıldığında, Resnet mimarisi derinliği arttıkca performansın olumsuz etkilendiğini Tablo \ref{tab:tab2} üzerinden görebilmekteyiz. 

\begin{table}[htbp]
\caption{DeeplabV3+ ile kullanılan farklı ResNet modelleri ile performans tablosu}
\begin{center}
\resizebox{.99\hsize}{!}{$
\begin{tabular}{|c|c|c|c|}
\hline
\textbf{Model} 	& \textbf{Öznitelik Çıkarıcı} 	& \textbf{mIoU}	& \textbf{Bina} 			\\ \hline 
DeepLabV3++    &  Resnet-18           	& \textbf{0.644}  &0.489          			             \\ \hline 
DeepLabV3++    &  Resnet-34           	& 0.631         	& \textbf{0.502}         	             \\ \hline 
DeepLabV3++    &  Resnet-50           	& 0.621         	&0.443          			     		\\ \hline 
\end{tabular}\label{tab:tab2}$}
\end{center}
\end{table}

Aynı zamanda yaptığımız bir diğer deneyde ise kayıp fonksiyonlarının etkisini inceledik. Bilindiği üzere Focal Tversky kayip fonksiyonu Dice kayıp fonksiyonunu kullanarak elde edilen bir kayıp fonksiyonudur. Bu kayıp fonksiyonunun Dice kaybından farkı  detaylı örnkeleri öğrenebilmek için \textit{gama} ve \textit{alpha} terimleri eklemiş olmakta ve bu geliştirmenin deneyde de elde ettirdiği iyileştirmeyi görmekteyiz. Burada deneylerde Focal Tversky kaybı içinde \textit{alpha} değeriyle oynanmış ve en iyi değer olarak 0.4 seçilmiştir. Fakat Dice kaybı kullanırken her sınıf için ayrı bir ağırlık verildiğinde bunun etkisinin en iyi olduğu gözlenmiştir. 

\begin{table}[htbp]
\caption{Bölütleme performansına Kayıp fonksiyonlarının etkisi}
\begin{center}
\resizebox{.99\hsize}{!}{$
\begin{tabular}{|c|c|c|c|c|}
\hline
\textbf{Model} 	& \textbf{Öznitelik Çıkarıcı} 	& \textbf{mIoU}	& \textbf{Bina IoU} 	& \textbf{Loss Function} 		 \\ \hline 
DeepLabV3++    &  Resnet-50           	& 0.473         	&0.166          			                    & 	Dice Kaybı			 \\ \hline 
DeepLabV3++    &  Resnet-50           	& 0.560         	&0.326          			                  & 	Focal Tversky Kaybı	 \\ \hline 
DeepLabV3++    &  Resnet-50           	& \textbf{0.621} &\textbf{0.443}         	         & Ağırlıklandırılmış Dice Kaybı	 \\ \hline 
\end{tabular}\label{tab:tab3}$}
\end{center}
\end{table}

\begin{figure}[htbp]
\centerline{\includegraphics[width=1.0\columnwidth]{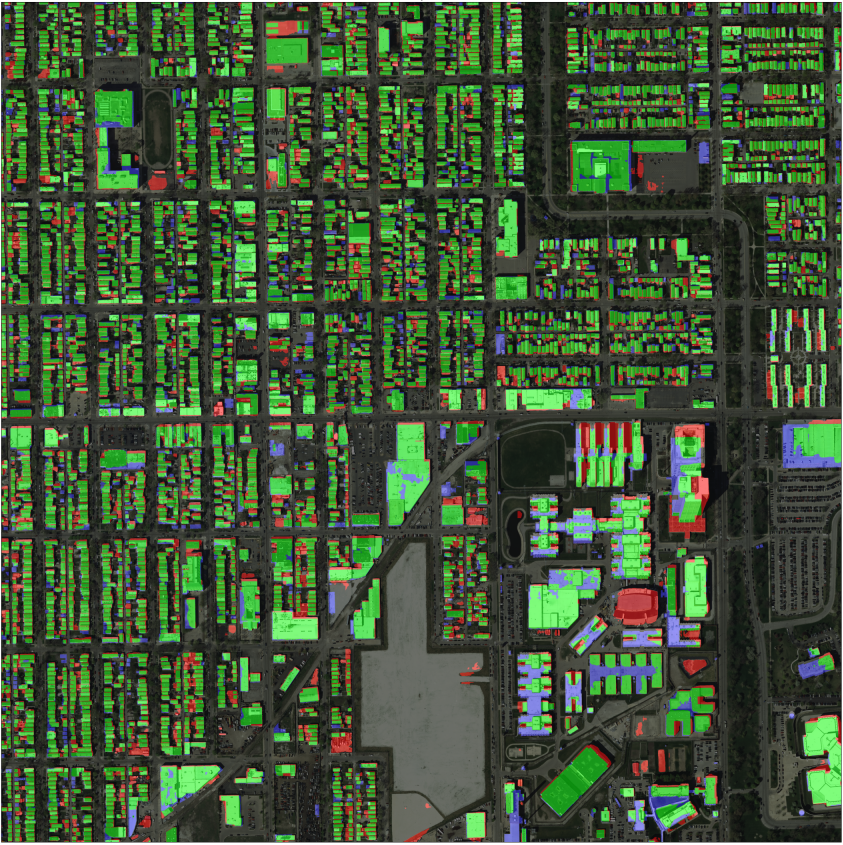}}
\caption{INRIA Chicago Veri kümesi 8. test görüntüsü tahmin sonucu}
\label{fig:overlayimage}
\end{figure}

Çalışmaların son kısmında model çıktılarının üzerinde test zamanında veri arttırımı yapılması performansı attırdığını gösteren çalışmalar olmuştur. Bu calışmada da birkaç farklı veri arttırımı yöntemi test zamanında kullanılmış ve test sonuçlarına etkisi gözlenmistir. 6 farklı deneme yapılmış olup bazılarının açıklaması tablo \ref{tab:tab5} de verilmiştir. \\

Tablo \ref{tab:tab5} da görülen metodlardan sadece Method-3 de \textit{Çarpma} veri arttırımı yapılmamış olup Method-1 ve Method-2 de yapılmıştır. Ayrıca sadece \textit{çözünürlük değiştirme} işlemlerinin yapıldığı test zamanı veri arttırımları da yapılmış olup bunların başarımı ölçülmüştür. Görüldüğü uzere \textit{çözünürlük değiştirme} yapılırken veri boyutunun arttırılması olumlu etki etmezken azaltıması olumlu etki etmektedir. Burada bir ayrı detay olarak \textit{Çarpma} veri arttırımı işleminin çıkarılması test zamanı veri arttırımı performansını arttırmıştır. Tablodan görülecegi üzere ilk baştaki sonuçlara göre 3\% ilerleme kaydedilmiştir. \\ 


\begin{table}[htbp]
\caption{Unet++ ve VGG-16 Öznitelik çıkarıcısı kullanılarak yapılmış son-işlem deneyleri}
\begin{center}
\resizebox{.99\hsize}{!}{$
\begin{tabular}{|c|c|c|c|}
\hline
\textbf{Veri Arttırım Yöntemi} 			& \textbf{mIoU}	& \textbf{Bina IoU} 	 & \textbf{Eğitim Veri Kümesi Çözünürlük Seviyeleri} 	 \\ \hline 
- 										& 0.701         	& 0.811          		 &  HWLC-18	(0.6m)				 \\ \hline 
Yöntem-1   								& 0.693         	& 0.809          		 &  HWLC-18	(0.6m)				 \\ \hline 
Çözünürlük Değiştirme [0.5 1]				& 0.714         	& 0.823          	       &  HWLC-18	(0.6m)				 \\ \hline 
Çözünürlük Değiştirme [0.25 0.5 1 1.25]	& 0.723         	& 0.831          		 &  HWLC-18	(0.6m)				 \\ \hline 
Çözünürlük Değiştirme [0.25 0.5 0.75 1]	& 0.724         	& 0.833          		 &  HWLC-18	(0.6m)				 \\ \hline 
Yöntem-2   								& 0.695         	& 0.809          		 &  HWLC-18	(0.6m)				 \\ \hline 
Yöntem-3   								&  \textbf{0.730}&  \textbf{0.836} 	 &  HWLC-18	(0.6m)				 \\ \hline 

\end{tabular}\label{tab:tab5}$}
\end{center}
\end{table}
\begin{figure}[htbp]
\centerline{\includegraphics[width=1.0\columnwidth]{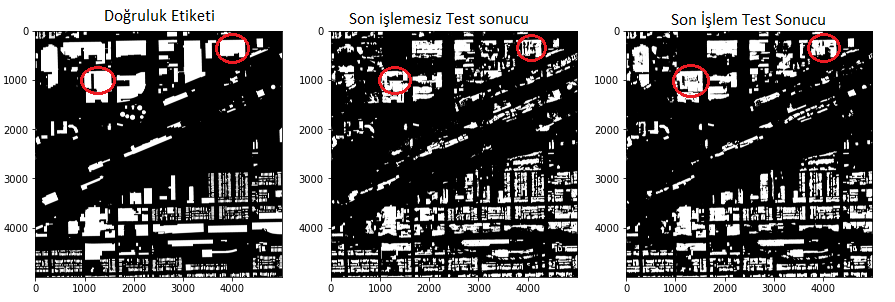}}
\caption{Son-İşlem yöntemlerinin Etkisi}
\label{fig:testtimeaugmentation}
\end{figure}

Yapılan deneylerde uydu görüntüsünün çözünürlüğünün hedef tespiti üzerinde etkisinin olduğunu görmekteyiz. Test görüntülerinin test zamanı sırasında çözünürlüğünün arttırılması, modelin eğitildiği çözünürlüğe yaklaştığı icin tespit performansı artmıştır. Bu sebeble yapılan deneylerde \textit{Çözünürlük Değiştirme} yapmak genel olarak modellerin test performansını arttıracağı öngörülmektedir. 

\section{Vargı}

Yapılan çalışmada 0.6 ve 2.4 metre çözünürlükteki HWLC-16 ve HWLC-18 verisetleriyle eğitilmiş modeller, INRIA veri kümesi Chicago bölgesinde test edilmiştir. Çalışmalarımızdan çıkardığımız sonuç itibariyle U-Net++ modeli VGG-16 öznitelik çıkarıcı ile en iyi sonuçları vermektedir. U-Net++ modelinin DeepLabV3 modeliyle yapılmış bütün test performanslarının üzerinde olduğu görülmektedir. Buradan çıkarılabilecek sonuçlardan biri DeepLabV3+ yönteminde yapılan delikli evrişim işleminin model derinliğini azaltması sebebiyle yeterince öğrenilemediğini ve özniteliklerin çıkarılamadığını düşündürmektedir.\\
Bunun yanı sıra, son-işlem yöntemleri ile test zamanında veri arttırımı sağlanarak çoklu test yapılmış, ve sonuçların \%3 seviyesinde artış göstermiştir. DeepLabV3 yöntemi üzerinde ResNet-18, ResNet34 ve ResNet-50 ile yapılan deneylerde Resnet-18 modelinin en iyi sonuç verdiği görülmüş, bunun sebebinin öznitelik çıkarıcılar arasında  alıcı alanı (receptive field) daha düşük olan versiyonun daha iyi performans verdiği görülmüştür. Keza VGG-16 yönteminin de alıcı alanı diğer EfficientNet ve ResNet mimarilerine göre daha düşük olduğu düşünülürse tutarlı olduğu görülmektedir. Bu çalışmada edinilen bir diğer önemli edinim, bina segmentasyonunda etkili kayıp fonksiyonu karşılaştırma deneyleridir. Bu deneylerde görüldüğü üzere ağırlıklandırılmış \textit{dice} kaybı diğer kayıp fonksiyonundan daha çok katkı sağlamıştır. \\
Uydu görüntülerinden bina ve obje tespitindeki en büyük zorluklardan biri farklı yakınlık seviyesindekindeki veri kümeleridir. Bu çalışmada hem Seviye-16 hem de Seviye-18 verileri ile eğitilmiş modeller yaklaşık olarak Seviye-15 yakınlığındaki farklı bir bölgede test edilmiştir. Modellerin performanslarında düşüş olmasına rağmen tatmin edici sonuçlar verdiği görülmektedir. Özellikle yakınlık seviyesi daha makul veri kümeleriyle eğitim yapmanın önemi buradan da görmüş bulunmaktayız. \\
Son olarak son-işlem testlerinde en faydalı yöntemin boyut arttırıp azaltmak olduğunu görülmüştür. Çözünürlük farklılıklarının da bulunması sebebiyle böyle durumlarda hem eğitim sırasında hem de sonrasında test görüntüsü çözünürlüğünün değiştirilmesi test sonuçlarına olumlu katkı sağlayabileceği düşünülmektedir.



\end{document}